\documentclass[runningheads]{llncs}
\usepackage[T1]{fontenc}
\usepackage{graphicx}
\usepackage[pagebackref,breaklinks,colorlinks]{hyperref}

\usepackage{color}

\usepackage{amsmath}
\usepackage{amssymb}
\usepackage{booktabs}
\usepackage{siunitx}
\usepackage{xspace}
\usepackage{tabularx} 

\usepackage[capitalize,noabbrev]{cleveref}

\def\ours{\textit{Cut-and-Splat}\xspace}

\begin{document}
\title{Cut-and-Splat: Leveraging Gaussian Splatting for Synthetic Data Generation}
\author{Bram Vanhele \and
Brent Zoomers \and
Jeroen Put \and
Frank Van Reeth \and
Nick Michiels
} 
\authorrunning{B. Vanherle et al.}
%
\institute{Hasselt University - Flanders Make - Digital Future Lab, Belgium\\
\email{firstname.lastname@uhasselt.be}}
\maketitle
\begin{abstract}
Generating synthetic images is a useful method for cheaply obtaining labeled data for training computer vision models. However, obtaining accurate 3D models of relevant objects is necessary, and the resulting images often have a gap in realism due to challenges in simulating lighting effects and camera artifacts. We propose using the novel view synthesis method called Gaussian Splatting to address these challenges. We have developed a synthetic data pipeline for generating high-quality context-aware instance segmentation training data for specific objects. This process is fully automated, requiring only a video of the target object. We train a Gaussian Splatting model of the target object and automatically extract the object from the video. Leveraging Gaussian Splatting, we then render the object on a random background image, and monocular depth estimation is employed to place the object in a believable pose. We introduce a novel dataset to validate our approach and show superior performance over other data generation approaches, such as Cut-and-Paste and Diffusion model-based generation.

\keywords{Synthetic Data \and Deep Learning \and Object Detection \and Instance Segmentation \and Gaussian Splatting}
\end{abstract}
\section{Introduction}
\label{sec:intro}

Deep neural networks are capable of solving complex computer vision problems. However, to do so, these models require a large number of annotated images specific to the problem they are solving. While obtaining numerous photos for a given problem is usually relatively straightforward, manually annotating these images is a very costly process. Certain annotation types, like semantic segmentation, can take humans dozens of minutes, while others, such as depth estimation and pose estimation, are complicated to do manually. Additionally, human-generated annotations may contain errors, biases, and inconsistencies, leading to a model performing poorly.

\begin{figure}
    \centering
    \includegraphics[width=0.8\columnwidth]{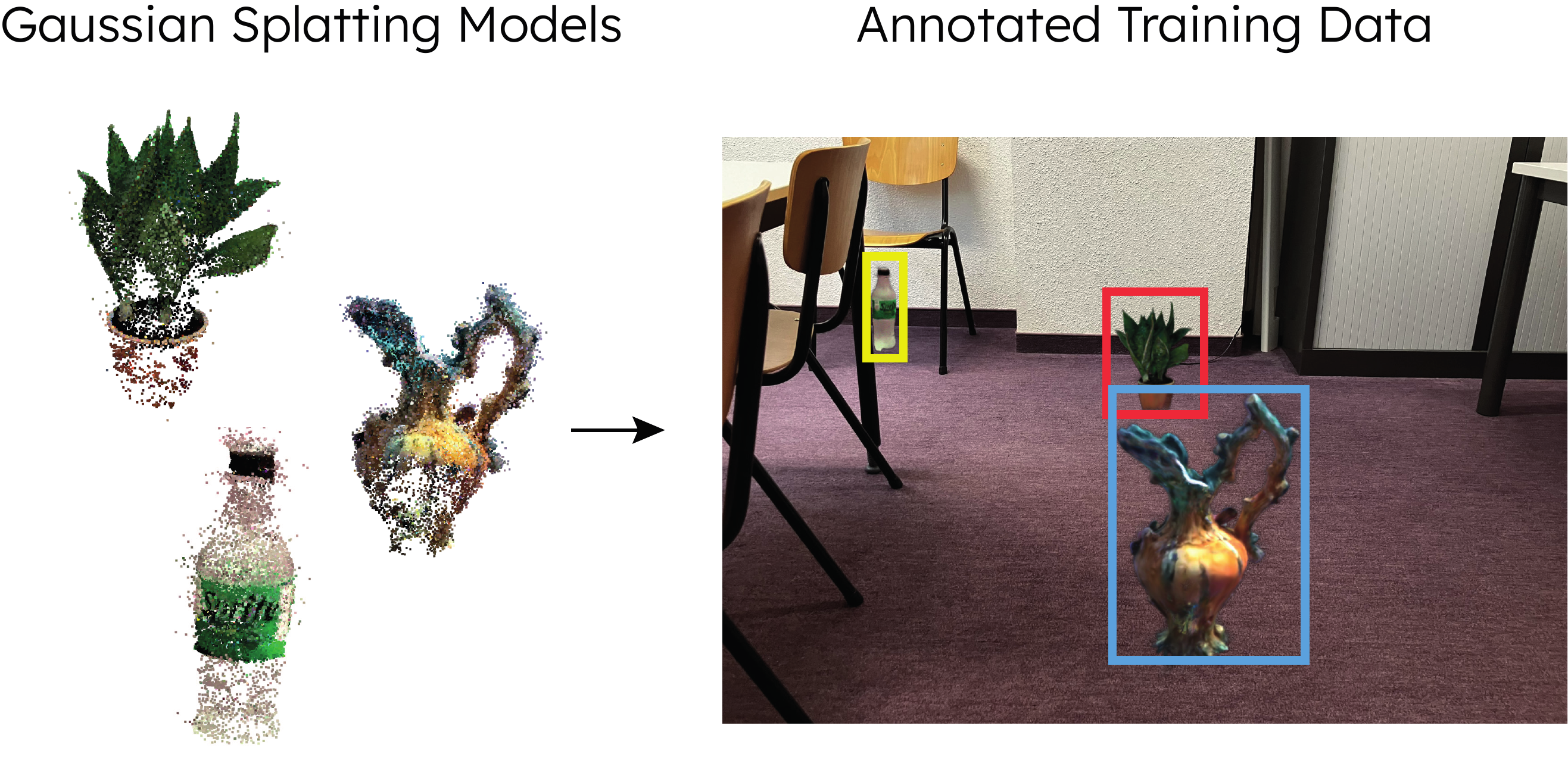}
    \caption{Our approach extracts foreground objects from trained Gaussian Splatting models and places them in plausible positions in background images to create high-quality synthetic images for training instance segmentation models.}
    \label{fig:overview}
\end{figure}

Using synthetic data alleviates some of these problems. Images are generated from a description of a scene. This description is then used to get the annotations for those images. The most common method for generating data is using 3D rendering engines such as Blender or Unity~\cite{greff2022kubric,moonen2023cad2render,borkman2021unity,wood2021fake}. A 3D scene composition containing the target objects is randomly created and rendered using a rendering technique such as rasterization or ray casting. This way, training data can be generated relatively quickly with pixel-perfect annotations. A downside of this technique is that the generated images can look quite different from the actual images due to the difficulty of accurately simulating light transport and camera sensor behavior. To keep this domain gap as small as possible, much information about the rendered scene is needed. Accurately textured 3D models are required for the target objects, and the target environment needs to be modeled as close as possible to the actual problem.

Synthetic data can also be generated using generative models, such as a GAN~\cite{li2022bigdatasetgan}, Diffusion Model~\cite{wu2023diffusion,azizi2023synthetic}, or using novel view synthesis techniques such as NeRF~\cite{ge2022neuralsim}. These learning-based approaches have the benefit that no textured 3D model is needed, and often, the domain gap is smaller as these models are trained to produce images based on the actual photographs. However, it is more difficult to control what is rendered since these techniques do not base their rendering on a 3D scene representation. Additionally, training data is needed to train these models, and artifacts can still be generated in the images.

This work focuses on creating synthetic datasets for detecting and segmenting specific objects in cases where a physical copy of the object is available for creating a dataset. The aim is to detect instances of that particular object or objects that are very similar. Hence, we do not consider very broad classes of objects such as \textit{dog}, \textit{tree}, or \textit{car}. This problem is more constrained but still has many valuable applications. Consider, for example, computer vision in a supermarket setting. They sell many different products, but all products of one type look largely the same. We strive to make the whole dataset creation process as convenient as possible by keeping the amount of manual work at a minimum. Our experiments focus on common household objects in natural settings.

We propose using the Gaussian Splatting~\cite{kerbl2023gaussians} technique for easy dataset creation. This novel view synthesis method learns to generate new viewpoints of a specific scene by optimizing the parameters of a set of 3D Gaussians. During training, these Gaussians' position, opacity, scale, rotation, and view-dependent color are optimized to represent the underlying scene as accurately as possible by minimizing a visual loss over the rendering of the training images. The technique lends itself well to creating synthetic data since only a short video of the target object is needed. Furthermore, the technique optimizes a 3D point cloud during training, which can be used to segment the foreground object. Additionally, each Gaussian has an opacity value, which allows for blending the extracted foreground object with background images.

Our proposed method, called \ours, uses the Gaussian Splatting method to generate context-aware synthetic data automatically. First, we capture a short video of a target object on a flat surface. We then train a Gaussian Splatting model for this object and automatically extract the Gaussians that make up the foreground object from the model. Next, we select a random background image and identify plausible support surfaces for the target objects in this image to ensure a logical scene composition. We use monocular depth estimation to find structure in the background image. The Gaussian Splatting model for the foreground object is then used to render it as if placed on that surface, resulting in a rendered image of the foreground object and an opacity map. This map is used to blend the foreground objects and background and to create object detection and instance segmentation annotations. The depth of the background is also used to ensure proper occlusions. You can see an example of some Gaussian Splatting models and an image created by our method in \cref{fig:overview}.

To evaluate our approach, we introduce a custom dataset specifically for evaluating image-based synthetic data generation approaches. Such a dataset is currently lacking in this field. It contains novel view synthesis input videos and two different validation sets taken from different cameras. Using this dataset, we perform an ablation that shows our approach can generate data that can serve to train good-performing instance segmentation models. We benchmarked our approach against other image-based data generation methods, such as Cut-and-Paste~\cite{dwibedi2017cut} and a Diffusion model~\cite{ho2020diffusion}. Our code and dataset are available at \url{github.com/EDM-Research/cut-and-splat}.

\section{Related Work}

Early forms of synthetic data generation use existing images of the target objects to generate new annotated images for object detection and instance segmentation. Cut, Paste, and Learn~\cite{dwibedi2017cut} uses a neural network to segment the foreground object from the existing images. We will refer to this work as Cut-and-Paste in this paper. These foregrounds are randomly transformed and placed on a random background image. Multiple blending modes are used so the network does not overfit on composition artifacts. Similarly, Ghiasi et al.~\cite{ghiasi2021simple} propose to generate additional samples by copying and pasting objects from one image to another using their existing segmentation masks. These methods allow for simple but effective data synthesis. Only a segmentation mask is needed for the foreground object, which can be done more efficiently using modern techniques such as Segment Anything~\cite{kirillov2023segany}. A downside of these approaches is that the generated images look unrealistic as implausible compositions are made, and artifacts could be introduced. Recent work has attempted to solve this by training a synthesizer network using a discriminator~\cite{tripathi2019compositing} or finding plausible locations to paste objects to~\cite{fang2019instaboost,dvornik2018modeling}. These methods are still limited by the existing viewpoints of the foreground objects, so they cannot be rendered in context correctly.

Another approach to synthetic data is using a rendering engine, such as Unity~\cite{moonen2023cad2render,borkman2021unity} or Blender~\cite{greff2022kubric}. Some solutions use less advanced rendering, such as OpenGL~\cite{hinterstoisser2019annotation}. Generating data this way gives complete control over all the parameters, making it a very flexible option for generating diverse training data. Objects can be rendered from all viewpoints. A significant downside is that textured 3D models are needed and that the often complex environments of the target data must also be modeled in 3D to achieve a small domain gap. Due to the difficulty of simulating the physical properties of light, it is challenging to render photorealistic images. This causes the domain gap to be more significant, which can negatively impact downstream performance. To overcome this, techniques such as Domain Randomization~\cite{tremblay2018domainrand} and Domain Adaptation~\cite{oza2023domainadaptation} have been applied.

This paper presents a synthetic data generation technique that includes the benefits of both approaches. We can generate foreground objects at all possible viewpoints by leveraging the novel view synthesis method of Gaussian Splatting, creating highly varied training data. A Gaussian Splatting model can be made from a video of a target object. Hence, no textured 3D model is needed. The novel view synthesis method can generate a highly accurate image representation of the object, leading to a smaller domain gap compared to render engines. Research has shown that object detectors use context when detecting and classifying objects~\cite{divvala2009context}. Our approach ensures plausible object context by finding logical surfaces in background images and leveraging novel view synthesis to correctly render the object in that position. When creating data with a graphics engine, the environment in which the objects are placed must also be modeled, which can add more complexity. Our approach can use any RGB image as an environment by using depth estimation to find structure.

PEGASUS~\cite{meyer2024pegasus} is another approach that leverages Gaussian Splatting for synthetic data generation. They focus on 6DoF pose estimation for robotics and use Gaussian Splatting models for the background environment as well. We differentiate by allowing any RGB image as a background, making it more convenient to introduce different environments in the datasets.

\begin{figure*}
    \centering
    \includegraphics[width=1\textwidth]{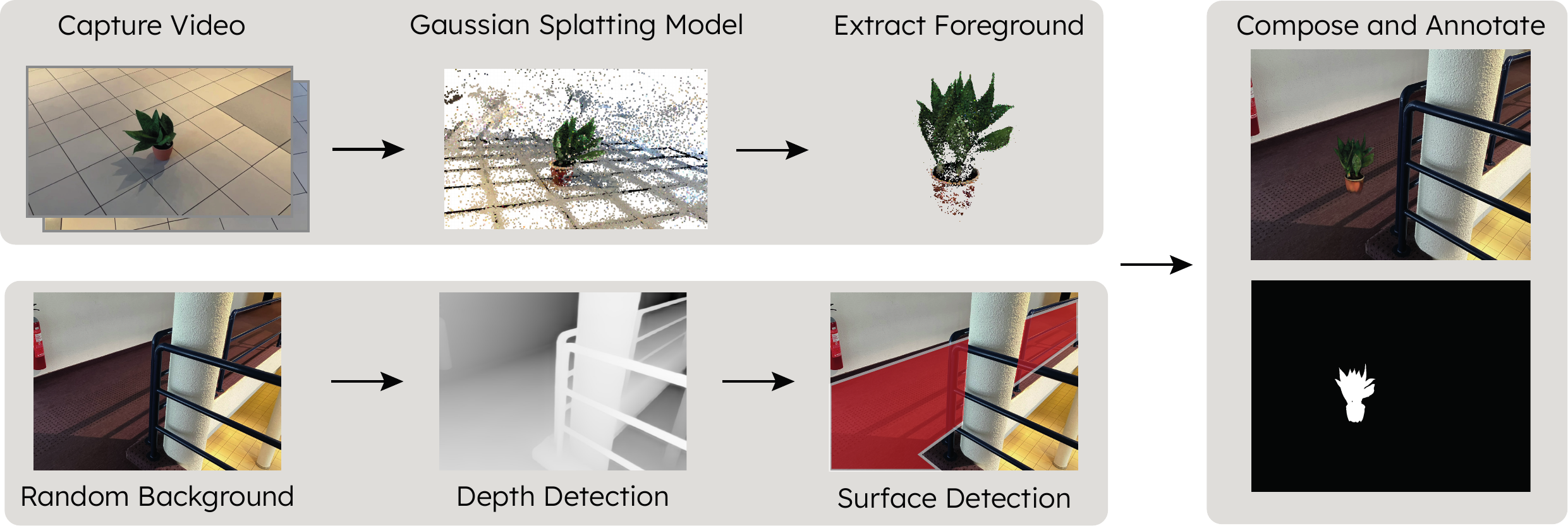}
    \caption{An overview of our method for easily creating realistic synthetic data. First, a Gaussian Splatting model is trained on a simple input video. The model representing the foreground is extracted. Second, an arbitrary background image is taken, and the depth is detected to find feasible placement positions. The Gaussian Splatting model is used to render the foreground object in a plausible pose.}
    \label{fig:method}
\end{figure*}

\section{Method}

In this section, we explain our method for creating a realistic and varied synthetic training dataset from a basic video of an object. The initial step involves training a Gaussian Splatting model for the target object. To train this model, a series of images of the target scene, along with their calibrated camera positions and an initial sparse point cloud, are required. These inputs can be easily obtained by recording a brief video of the object and using a structure-from-motion algorithm such as COLMAP~\cite{schonberger2016sfm}. A video of about one minute is sufficient, and this is the only manual action required to create a dataset with our approach.

The Gaussian Splatting models trained for each object still contain Gaussians representing the background scene, which do not belong to the target object. This is undesirable, as we only want to generate training data containing the foreground object. Therefore, we automatically extract the Gaussians relevant to the target object from the complete model. A training image for the downstream task is created by selecting a random image from a set of background images and rendering the foreground object in this image. This is done by finding support surfaces in the background image and ensuring the correct perspective for the foreground objects to create realistic images. Realism is further improved by taking depth into account. \cref{fig:method} shows an overview of our proposed method.

\subsection{Foreground Object Extraction}

After optimization, the object is defined by a set of 3D Gaussians. Each Gaussian is centered around a mean $\mu$, representing a location in 3D space. To separate the foreground object from the trained model, we make the assumption that the object is situated on a flat surface, like the floor or a table. This assumption enables us to identify the foreground object by filtering out the ground plane. We apply Random Sample Consensus (RANSAC)~\cite{fischler1981ransac} to the point cloud defined by the $\mu$ values. Specifically, we select three random points from the point cloud and count the number of points that lie close to the plane defined by those three points. This is repeated for several iterations, and the plane with the most inliers is considered the ground plane. This results in a set of points that belong to the ground plane.

The Gaussians corresponding to the points on the plane are removed from the model. The internal point cloud of the Gaussian Splatting model is not geometrically perfect. Some points belonging to the ground plane are thus not marked as such since they deviate too much from the detected plane. These points are very sparse, so we can filter them out using a statistical filter that removes points further away from their neighbors compared to the average for the point cloud. For this, we consider 50 neighbors and keep points less than 0.1 standard deviations away from their neighbors. Decreasing the ratio makes the filter more aggressive and will remove more noise. The geometry of the target object is robust to this filter, as it has many dense points since it is the focus of the Gaussian Splatting model.

\begin{figure}
    \centering
    \includegraphics[width=1\columnwidth]{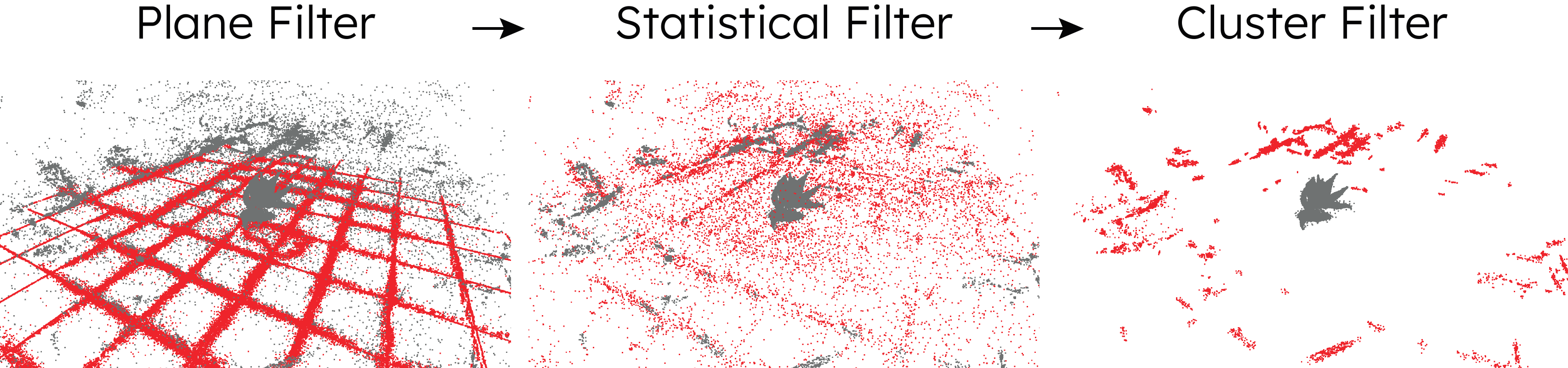}
    \caption{A subsequent plane filter, statistical filter, and cluster filter are used to extract the plant object from the point cloud representation of the trained Gaussian Splatting model. Red illustrates points that are selected for removal.}
    \label{fig:filtering}
\end{figure}

After separating the foreground object from the resting plane, a halo of points around the target object that the statistical filtering algorithm did not successfully remove can remain. These background points exist because they are outside of the focus of the input video. Hence, there are not many observations of these points, and the Gaussian Splatting representation is noisy. We filter out these background points by applying the DBSCAN~\cite{ester1996dbscan} clustering algorithm to the point cloud. Assuming that the target object is roughly at the center of the point cloud, we keep the cluster closest to the middle. The middle is the average of all points. For DBSCAN, we use an $\epsilon$ value of 0.5 and a minimum of 100 points. Since the previous filtering step created a large gap between the target object and the halo of noise, the parameters of this step are not sensitive. \cref{fig:filtering} contains an illustration of the three filtering steps that are done to extract a foreground object.

\subsection{Object Placement}

Existing synthetic data generators often paste the object over a random selection of background images with no regard for the physical plausibility of the resulting scene configuration. This can result in awkward images of objects completely out of place, objects floating in the air, or objects scaled too large or small to fit into the background scene. Previous methods do not consider the background image's perspective and ignore that objects are most often in a resting position when photographed. Research has shown that context can be important when training object detection models~\cite{dvornik2018modeling}.

Therefore, we make sure that the foreground objects are placed in more realistic scene configurations. We achieve this by using the monocular depth estimation technique called Depth Anything~\cite{yang2024depthanything} to generate depth maps for the background images. This approach allows us to use any collection of RGB images from the internet as backgrounds. It also enables users to capture relevant images for their specific problem domain. As a result, the synthetic data produced more closely matches the real-world conditions, bridging the domain gap even further.

The depth map of the background image is converted to a point cloud using an estimation of the intrinsic matrix by assuming an FOV of 55 degrees and a central principal point. While not perfectly accurate, this leads to visually good results. Statistical outliers are removed from this point cloud.

\begin{figure}
    \centering
    \includegraphics[width=\linewidth]{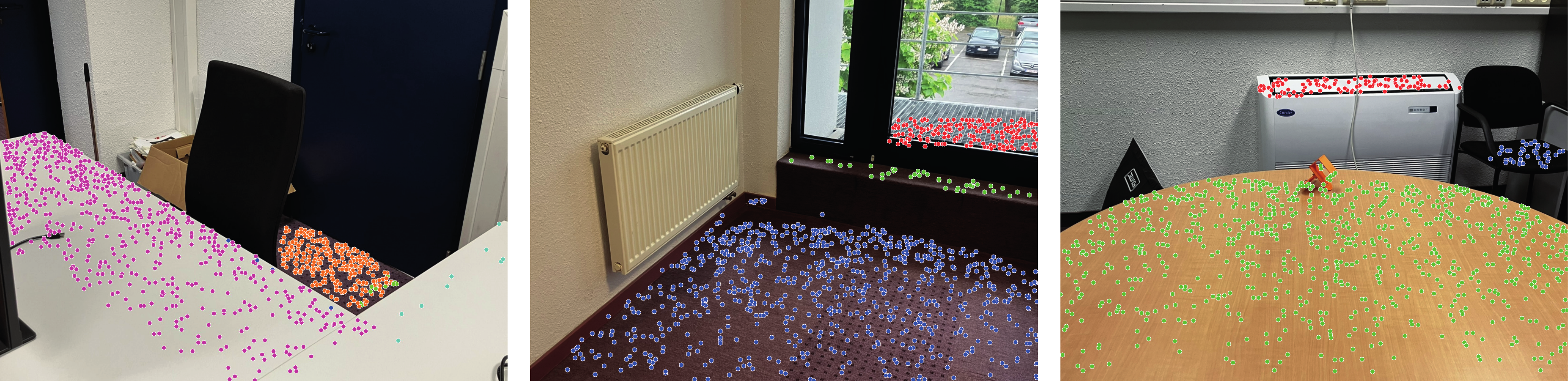}
    \caption{Illustration of possible object placements in the background images computed by our approach. For each image, we show 1000 possible placement positions, indicated by colored dots. A different color indicates a different plane.}
    \label{fig:placement}
\end{figure}

We take the practical assumption that most of the resting surfaces can be approximated as a plane in the background scene, e.g., a table, chair, or floor. To find multiple planar surfaces in the point cloud, we use a statistics-based algorithm~\cite{aruajo2020planes}. We keep only surfaces that are roughly horizontal with respect to the orientation of the scene. We find the up-axis of the scene based on the PCA~\cite{pearson1901pca} of the convex hull of the point cloud. Based on this up-axis, we filter out planes that are not horizontal to avoid placement on, for example, walls. Additionally, horizontal planes that are too close to the top of the scene are also removed. When picking a random plane to place an object on, the probability each plane is selected is based on its surface area. For computations on point clouds, we use Open3D~\cite{zhou2018open3D}. We assume resting surfaces are horizontal, but this assumption can easily be relaxed. The normals of the detected planes will later be used to align the target object to the surface. \cref{fig:placement} illustrates some potential object placement positions found by our approach.

\subsection{Scene Composition}

To generate a new scene composition, we select a foreground object and one of the approximately horizontal surfaces in the background to place it on. A random point on that surface is chosen as a final destination for the object. We position the foreground object in the scene by taking into account the surface normal of the fitted plane and rotating the object to its upright resting position. The resting configuration for objects is chosen by aligning its normal in the foreground point cloud data with the surface normal of the resting plane. The normal of the foreground object is the normal of the filtered-out plane from the Gaussian Splatting model. This process ensures depth-dependent scaling and perspective-correct placement. To increase variation in the representations of the object in the dataset, we rotate the object randomly around its local up axis. There is no mechanism that ensures correct inter-object scaling.

\begin{figure}
    \centering
    \includegraphics[width=0.8\linewidth]{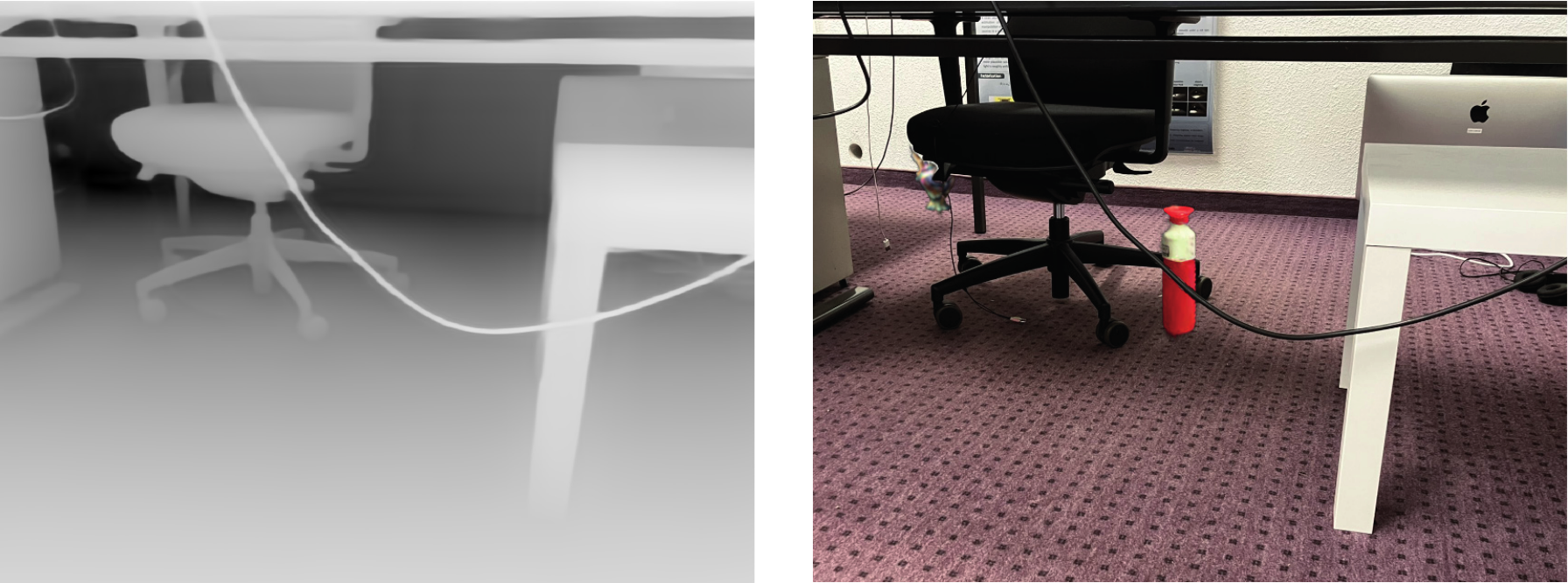}
    \caption{The red bottle is rendered on the floor in the background (right). The depth map (left) computed by Depth Anything is used to realistically occlude the object behind the cable.}
    \label{fig:occlusion}
\end{figure}

Once the foreground object's position, orientation, and scaling are determined, the Gaussian Splatting model is evaluated with those parameters, and the foreground object is rendered from the appropriate viewpoint. The rendering is done once to obtain the color values and then again with the splat color set to white to obtain the opacity map. Next, the opacity values are filtered by considering the background scene's depth values to have realistic occlusions. A median filter is applied to the background depth map to prevent noisy occlusions. \cref{fig:occlusion} shows an example of how the depth is used to simulate occlusion. The final opacity map is then used to blend the foreground object realistically into the background scene.

\begin{figure}
    \centering
    \includegraphics[width=0.8\columnwidth]{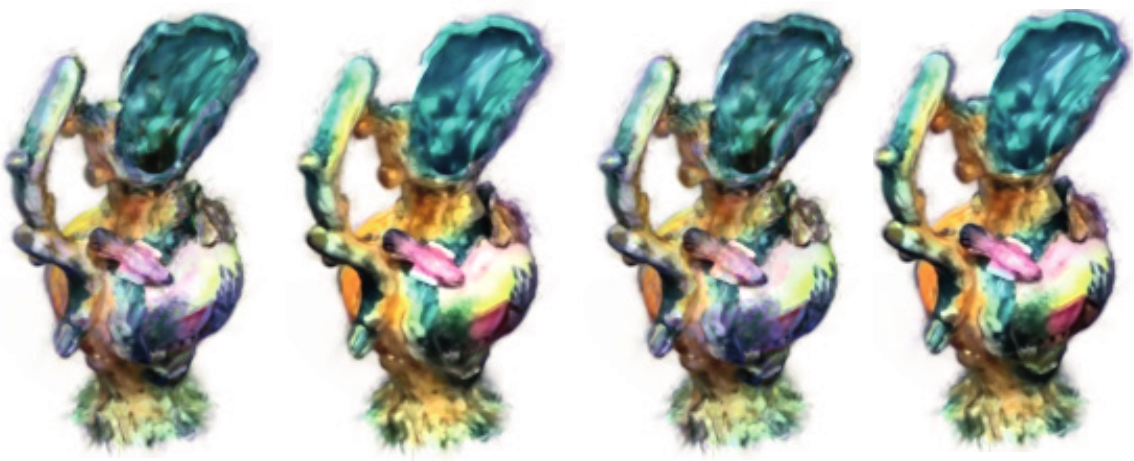}
    \caption{By varying the angle at which the spherical harmonics are evaluated with respect to the camera, we introduce subtle variations.}
    \label{fig:augmentation}
\end{figure}

The lighting is not adapted to the background scene when placing the Gaussian Splatting renders in the background images. Due to this, the object's appearance is limited to the lighting conditions from the captured video. This could cause the model to overfit this visual representation of the object and not generalize to other lighting conditions. To avoid this, we augment the object's appearance in two ways. Firstly, when rendering the Gaussians for the foreground, we use a random vector to calculate the spherical harmonics instead of the actual camera position. This changes the object's appearance even when the camera angle remains the same, introducing more variations. This is illustrated in \cref{fig:augmentation}. Additionally, we apply pixel-level augmentations such as blurring, color adjustments, noise addition, and random tone curves. While these augmentations may not simulate realistic lighting, they introduce additional variation to help the model adapt to different lighting conditions in the test data.

After rendering, the opacity maps of the foreground objects are used to export annotations that can be used to train detection models on the created images. These annotations are bounding boxes and instance segmentation masks.

\section{Results}

In this section, we demonstrate that our approach can efficiently generate datasets suitable for training high-performing instance segmentation models. We use our method to create multiple datasets for training such models and evaluate their performance using real annotated data. To show the individual importance of the components of our method, we perform an ablation. Next, we compare our approach to two other image-based synthetic data generation approaches. Cut-and-Paste~\cite{dwibedi2017cut} is an approach that is often used due to its effectiveness and simplicity. Since Diffusion models~\cite{ho2020diffusion} have the ability to generate very realistic-looking images, we also include an experiment comparing our approach to synthetic data generated by a Diffusion model. We do not compare our approach to NeRF-based data generation since the foreground object is more difficult to extract from these models. This could have a large impact on the results.

In all experiments, we train a Mask R-CNN~\cite{he2017mask} model with a ResNet50~\cite{he2015resnet} backbone. The model is trained for 100 epochs, with 1000 images for each epoch. The model backbone is initialized with weights trained on ImageNet~\cite{deng2009imagenet}, and some basic image augmentations are used during training. We compute the mean average precision (mAP) over the 0.5 to 0.95 overlap threshold range as a validation metric. This computation is done over the predicted bounding boxes. When training Gaussian Splatting models, we use the standard Gaussian Splatting version and implementation~\cite{kerbl2023gaussians} with all the default parameters.

\subsection{IBSYD Dataset}

\begin{figure}
    \centering
    \includegraphics[width=0.8\linewidth]{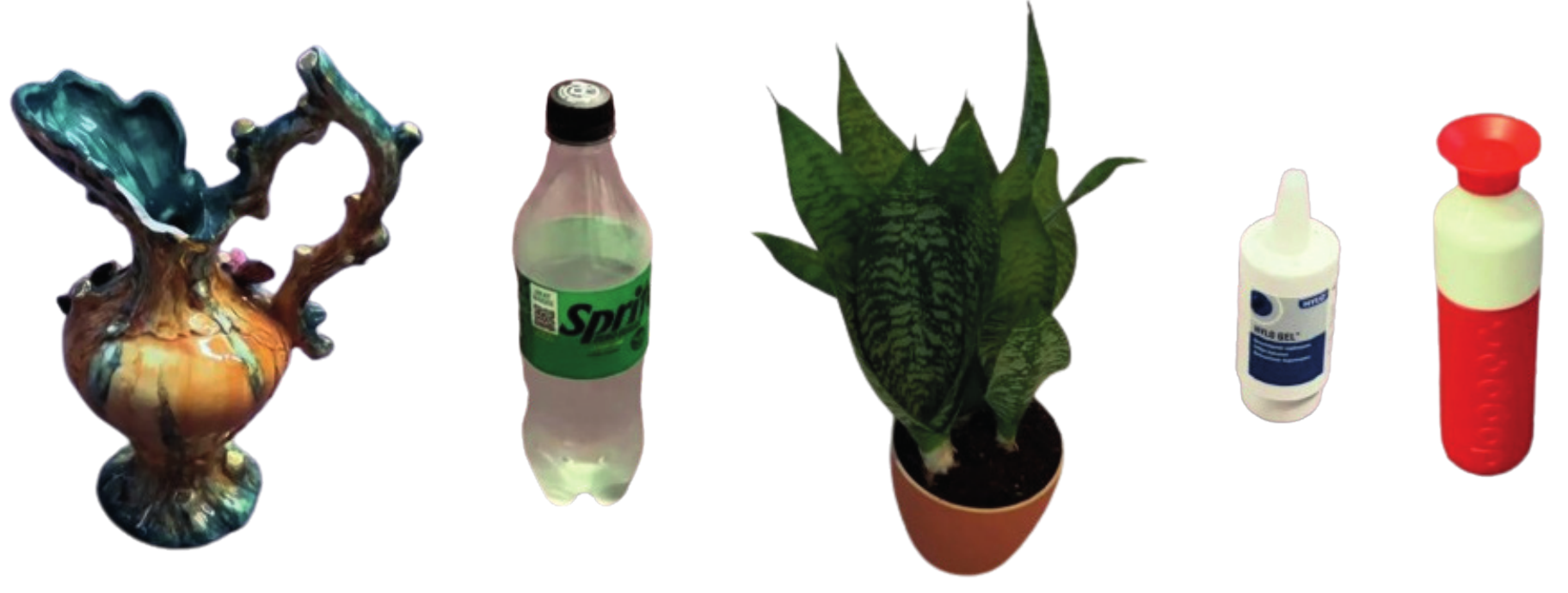}
    \caption{We introduce the Image Based Synthetic Data (IBSYD) dataset, which contains five challenging and varied objects.}
    \label{fig:objects}
\end{figure}

Leveraging novel view synthesis for synthetic data generation is a relatively new domain. To thoroughly test our approach, we, therefore, introduce a custom dataset. The \textit{Image Based Synthetic Data (IBSYD)} dataset. The dataset describes several challenging and diverse objects: a bottle of eyedrops, a plant, a semi-transparent bottle of soda, a water bottle, and a colorful vase with a handle. These objects are illustrated in \cref{fig:objects}. For each object, a video is provided that can serve as input for a novel view synthesis method such as ours. Each object was placed on the floor separately, and a one-minute video was recorded with a smartphone. This process takes only a few minutes in total and is the only manual action needed.

\begin{figure}
    \centering
    \includegraphics[width=0.8\linewidth]{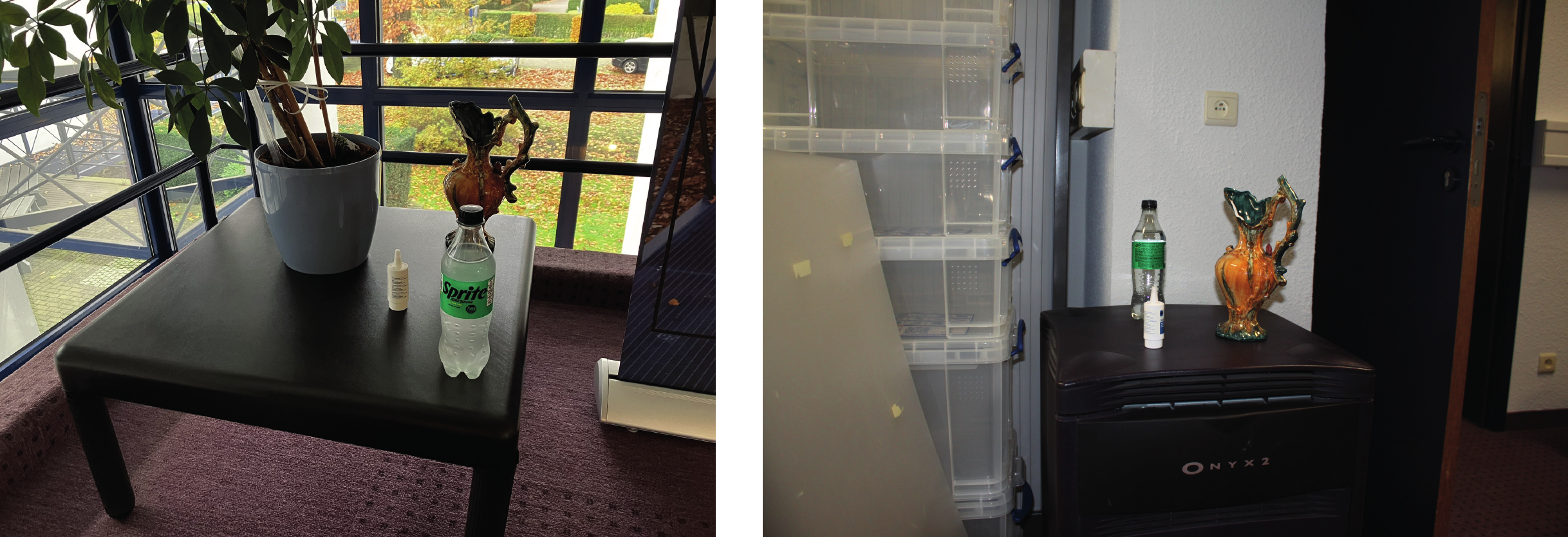}
    \caption{The IBSYD dataset contains two validation datasets. One taken with an iPhone camera (example on the left) and one taken with a Canon camera (example on the right).}
    \label{fig:ibsyd_validation}
\end{figure}

Additionally, the dataset includes a test set of real images to validate whether the generated images can be used to train detection models. We manually took several photographs, each containing one to three objects. The objects are placed naturally, i.e., on a tabletop or the floor, and have occlusions. Photographs are taken from a wide variety of indoor scenes. We used two different cameras to create these photographs. 50 photographs were taken with the same iPhone used to create the input videos, and 50 were taken with a Canon 500D camera. This allows us to investigate if the generated datasets are overfitted to the camera used to train the Gaussian Splatting model. The images from the Canon camera differ significantly from those from the smartphone as the camera has a different lens and sensor, the objects are sometimes out of focus, and in some cases, the flash was used. The distribution of object occurrences and combinations is the same between the two validation sets. \cref{fig:ibsyd_validation} shows an example from both datasets.

\subsection{Ablation}

\begin{figure}
    \centering
    \includegraphics[width=0.8\linewidth]{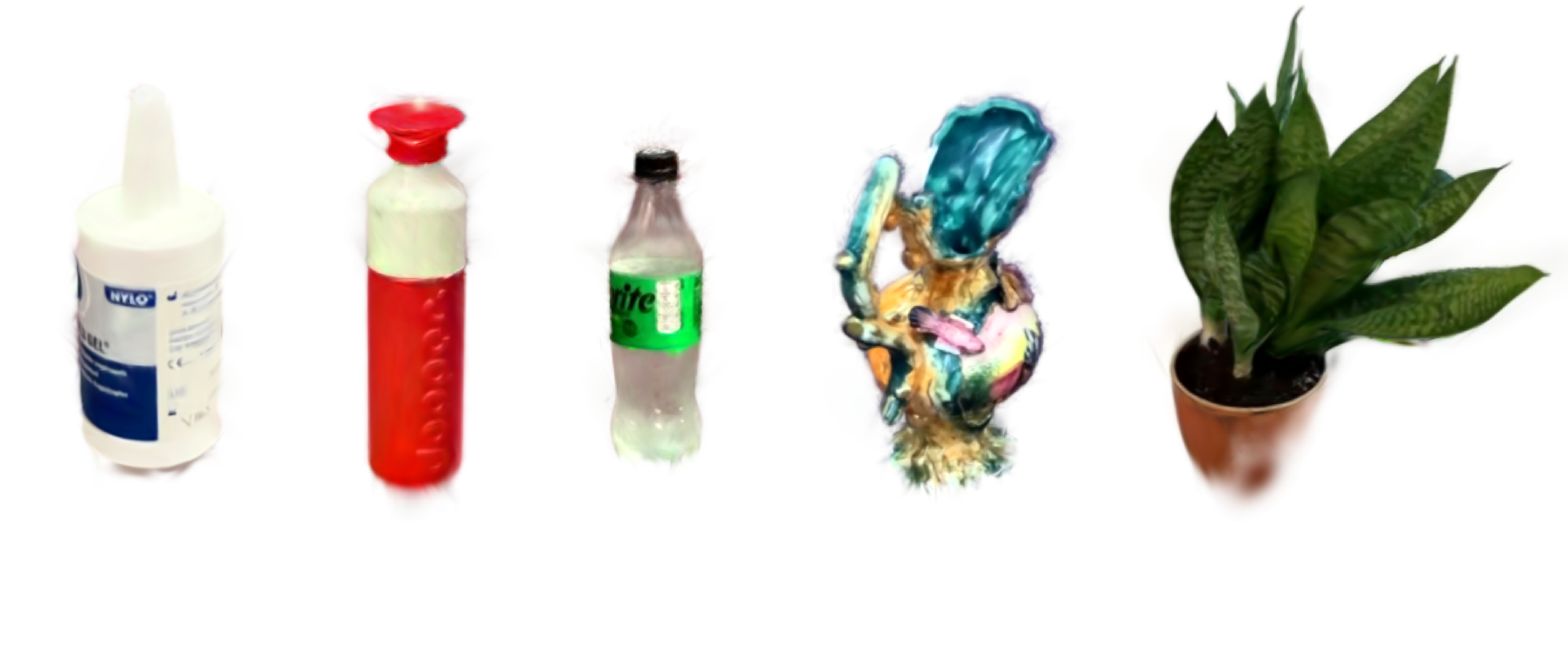}
    \caption{Objects rendered by Gaussian Splatting after segmenting the foreground objects.}
    \label{fig:renderd}
\end{figure}

\begin{figure}
    \centering
    \includegraphics[width=1\linewidth]{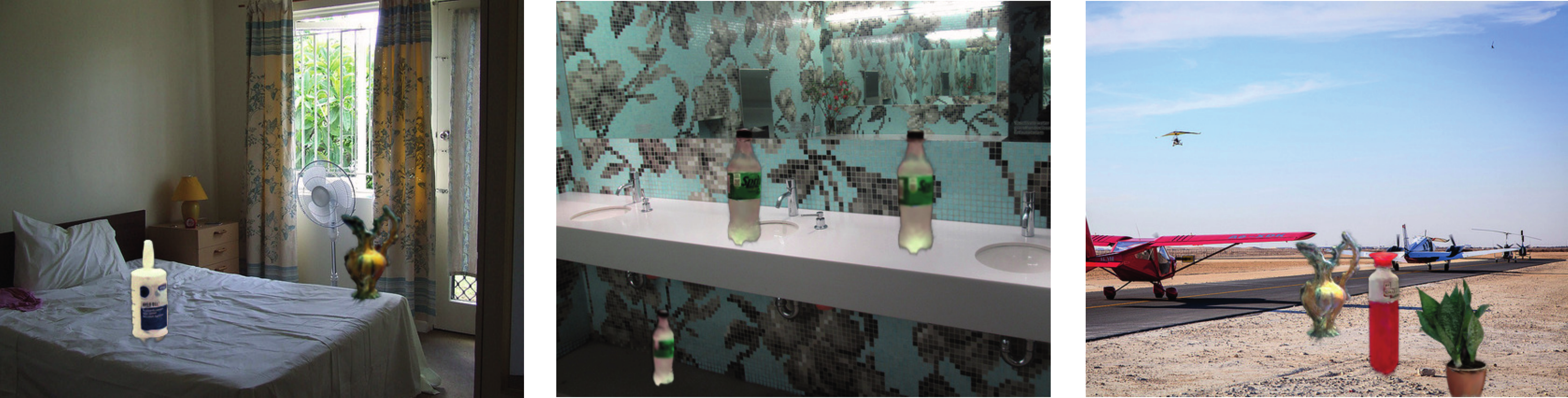}
    \caption{Some examples of images rendered by our approach.}
    \label{fig:examples_ours}
\end{figure}

First, we perform an ablation to highlight the abilities of our approach and to show the importance of the different components proposed in our pipeline. Namely, we investigate the impact of the smart placement technique and augmentation of the representation of the foreground object. For this reason, we generate three different datasets with our \ours approach using the input videos of the aforementioned IBSYD dataset. One using the full approach, one without augmentation, and one without smart placement. For the latter, objects are rendered from a random camera angle and placed in a random position on the screen. We train an instance segmentation model on each of these datasets and test it on the two different test sets. The background images are taken from the COCO dataset~\cite{lin2014coco}. Each rendered dataset contains \num{5000} images, and each image has one to three objects. Some visualizations of the Gaussian Splatting models trained for our objects are shown in \cref{fig:renderd}.\cref{fig:examples_ours} shows several images created by our dataset. We observe that objects are placed on surfaces with plausible poses.

\begin{table}[]
    \centering
    \caption{Performance of instance segmentation models trained on datasets generated by different variations of our method.}
    \label{tab:ablation}
    \begin{tabularx}{0.7\textwidth}{Xcc}
    \toprule
    \textbf{Dataset}       & \textbf{$\text{mAP}_{\text{iPhone}}$} & \textbf{$\text{mAP}_{\text{Canon}}$} \\
    \midrule
    \textbf{\ours} & \textbf{81.21}                          & \textbf{79.68}                         \\
    no augmentation        & 71.17                                   & 73.71                                  \\
    no smart placement     & 52.05                                   & 53.51                                   \\
    \bottomrule
    \end{tabularx}
\end{table}

\begin{figure}
    \centering
    \includegraphics[width=\linewidth]{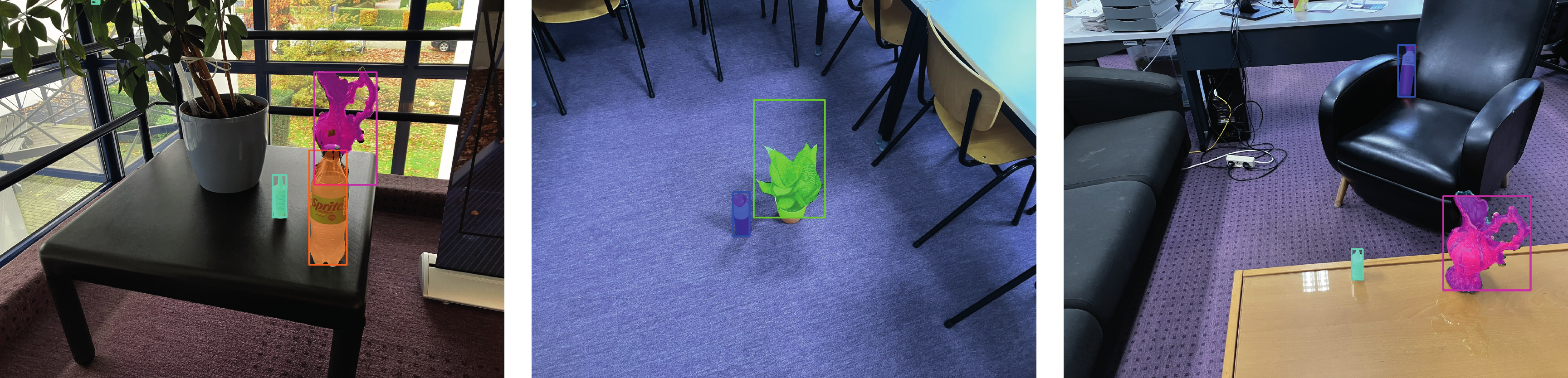}
    \caption{Some successful detections on the iPhone test set made by an instance segmentation model trained on a dataset generated by \ours.}
    \label{fig:detections}
\end{figure}

The results in \cref{tab:ablation} show that a model trained on a dataset generated by \ours delivers very good results, scoring around 80 mAP. This holds true for both the iPhone and Canon validation datasets. This indicates that the trained model does not overfit on the camera used to capture the Gaussian Splatting model. Furthermore, we observe a significant decrease in performance when no augmentation is used when rendering the foreground images. This highlights its importance in overcoming the lack of realistic lighting on the rendered objects. Finally, when objects are rendered in random positions, we see a very large drop in performance to almost 50 mAP. This shows the benefit of realistically placing objects in the background scenes. Thanks to this, the training data is more similar to the target domain, leading to a large increase in performance. Some successful detections made by the model trained on the full \ours dataset are shown in \cref{fig:detections}.

\subsection{Versus Cut-and-Paste}

In the Cut-and-Paste approach, foreground objects are cropped from a limited set of input images and placed on background images following a random transformation. The downsides are that only a limited amount of viewpoints are considered and that objects are placed without considering context. However, this approach does not suffer from artifacts introduced by Gaussian Splatting like our approach.

\begin{figure}
    \centering
    \includegraphics[width=\linewidth]{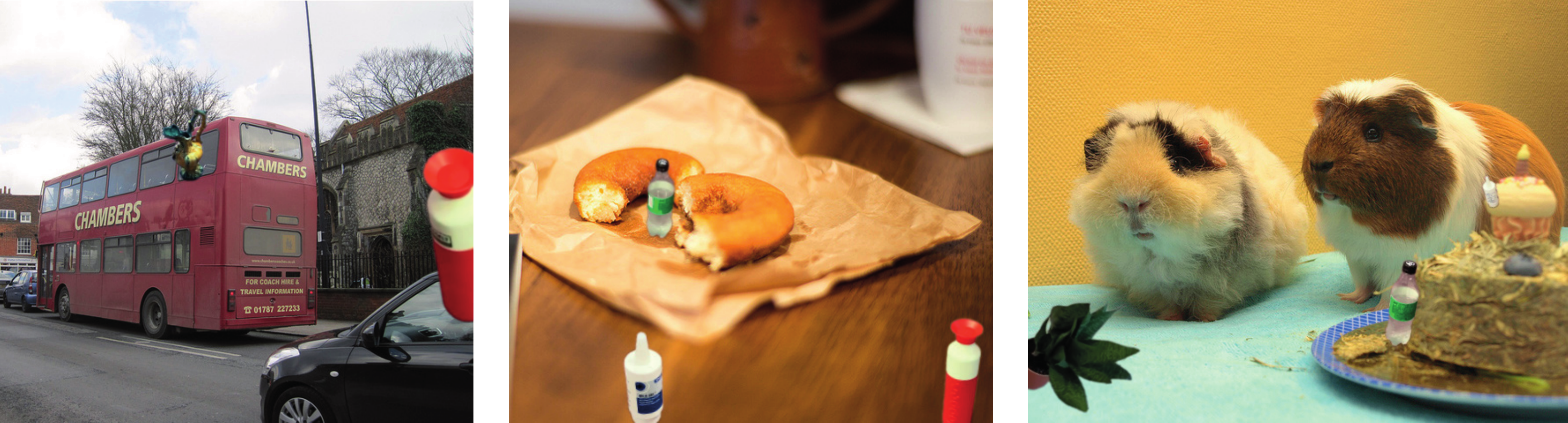}
    \caption{Some examples of images generated using the Cut-and-Paste approach.}
    \label{fig:cutandpaste}
\end{figure}

We create a Cut-and-Paste dataset of the IBSYD dataset following the same specifications as in our previous experiment, i.e., COCO backgrounds, 5000 images, and up to three objects per photo. For each object, we select ten frames from the input video and manually extract the foreground mask using Segment Anything~\cite{kirillov2023segany}. The frames are taken from widely varying camera positions to ensure a good representation of the object. Some examples of this dataset are shown in \cref{fig:cutandpaste}.

The COCO images we used as backgrounds for generating data do not match the target domain well. This often leads to strange-looking images, even when we use our method for realistic scene composition. For this reason, we introduce a set of relevant background images for this experiment. We took 80 photographs in the same rooms where the test set was recorded. The objects of the test set do not occur in these photographs. The same camera as in the iPhone test set is used, and the photographs are taken from random poses that do not match the test set. Using these backgrounds, we create a second dataset using the Cut-and-Paste approach. As a comparison, we also create a \ours dataset using these backgrounds.

For both our approach and the Cut-and-Paste method, we create two datasets. One with unrelated background images and one with in-domain background images. An instance segmentation model is trained on each of these datasets and tested on the iPhone test set. This way, we compare the performance of \ours to that of Cut-and-Paste approach, and we research whether using in-domain backgrounds leads to higher-quality synthetic data

\begin{figure}
    \centering
    \includegraphics[width=0.8\linewidth]{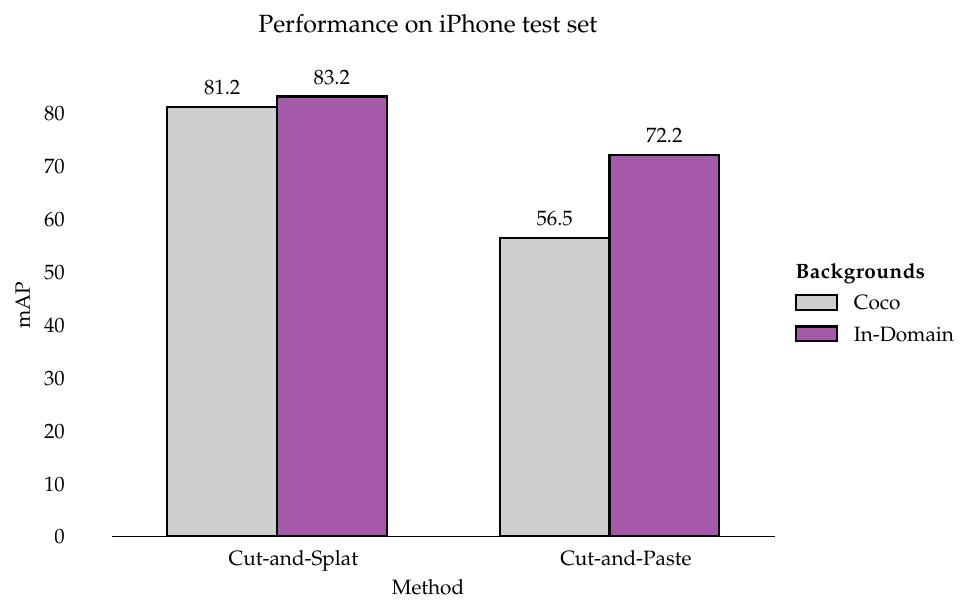}
    \caption{The performance of instance segmentation models trained on data generated by \ours versus Cut-and-Paste. We consider using both unrelated COCO backgrounds and domain-relevant background images.}
    \label{fig:vs_cutandpaste}
\end{figure}

The results in \cref{fig:vs_cutandpaste} show that models trained on datasets generated by our approach outperform those trained on Cut-and-Paste datasets. This shows that the plausible object context and increase in viewpoint variation in our method leads to better datasets. We observe that for \ours, the in-domain backgrounds lead to a small increase in performance. For Cut-and-Paste, using in-domain backgrounds gives a much larger boost in performance. This indicates that our approach is more robust to out-of-domain backgrounds. Hence, less effort needs to be spent collecting background data.

\subsection{Versus Diffusion Model}

There are multiple different ways to use a Diffusion model to generate synthetic data. To avoid training or fine-tuning a Diffusion model, we use an approach similar to Cut-and-Paste. We randomly place the foreground objects on an image and use Stable Diffusion XL~\cite{podell2024sdxl} inpainting to generate a background. This has the benefit that the background is generated to fit the foreground images to create a realistic-looking image. Additionally, the foreground images are slightly adapted to the generated background, creating even more realism. We use ChatGPT to generate random prompts that ask the Diffusion model to place the objects in indoor environments.

\begin{figure}
    \centering
    \includegraphics[width=0.30\linewidth]{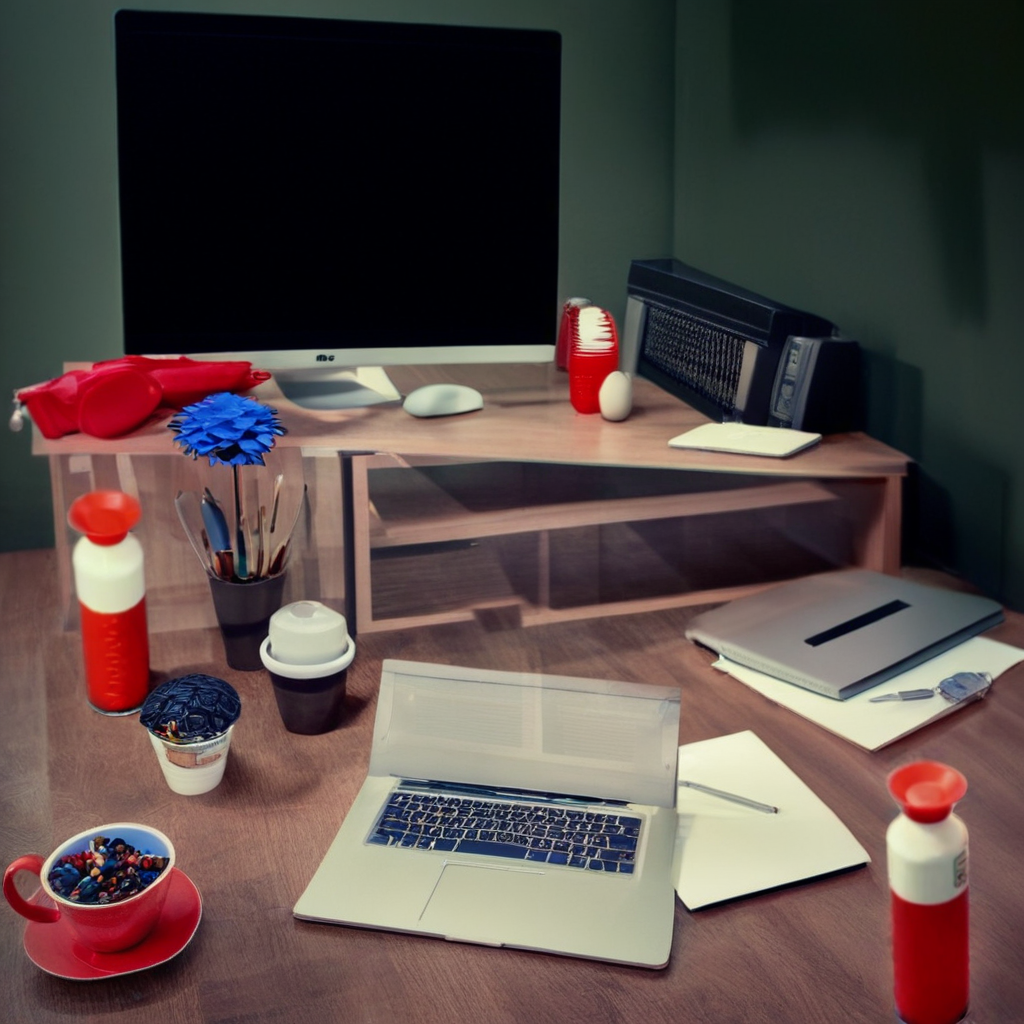}
    \includegraphics[width=0.30\linewidth]{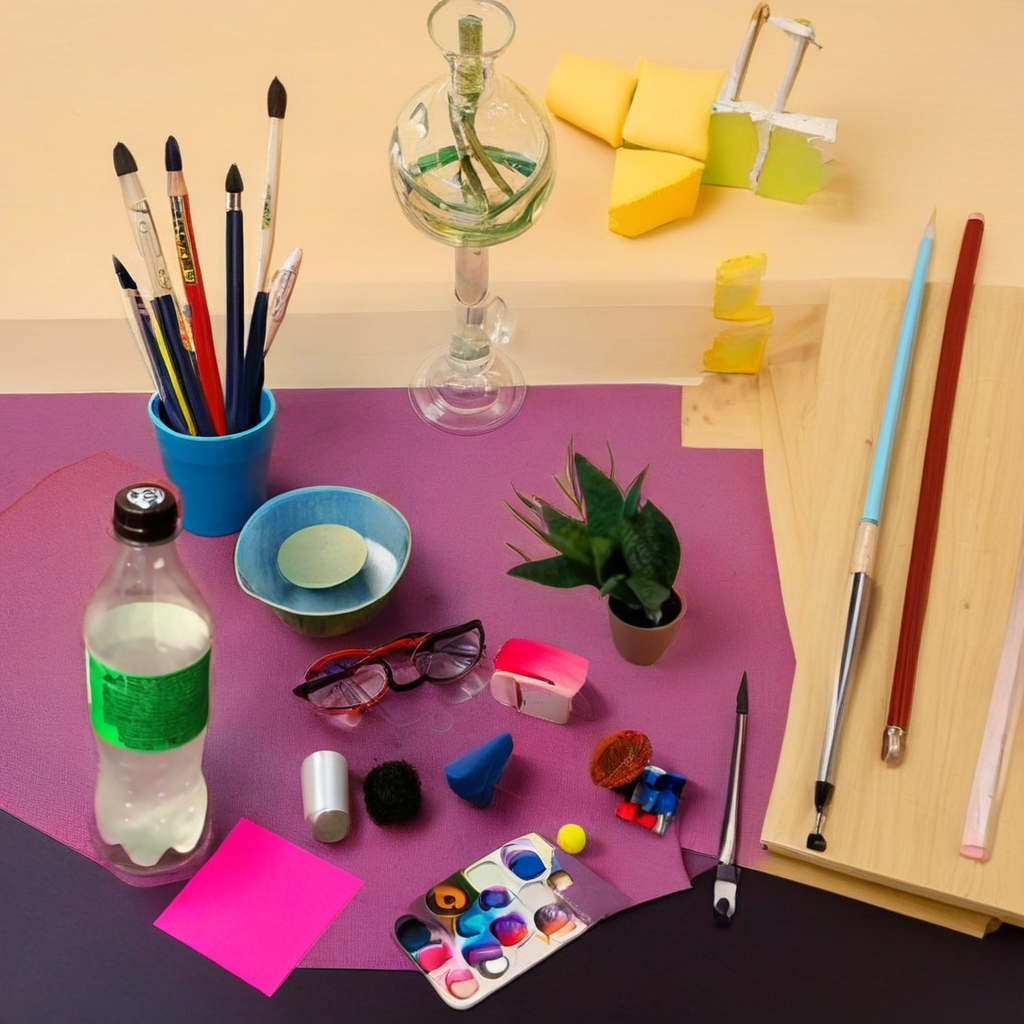}
    \includegraphics[width=0.30\linewidth]{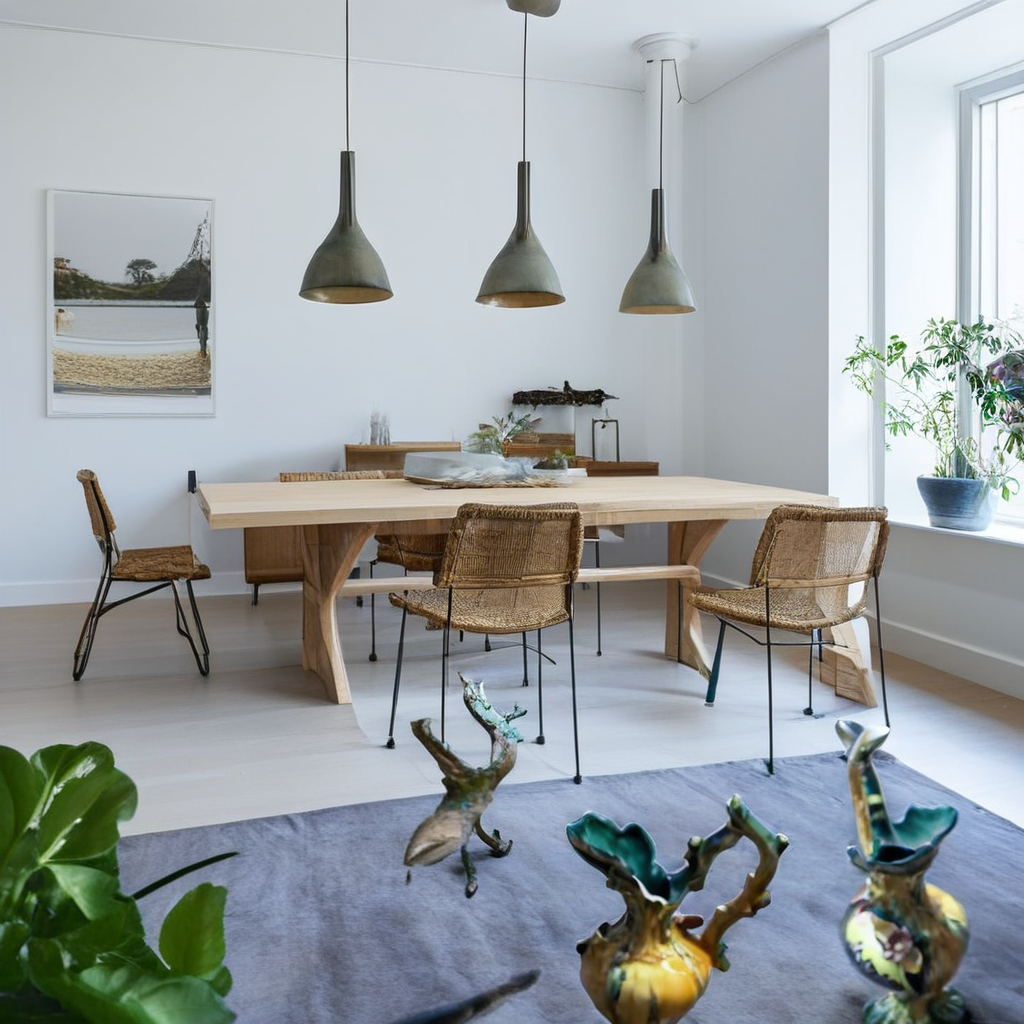}
    \caption{Some examples of images generated using the Diffusion model approach.}
    \label{fig:diffusion_samples}
\end{figure}

Using this approach, we generate a 5000-image synthetic dataset for the IBSYD objects. Some samples of this dataset are shown in \cref{fig:diffusion_samples}. An instance segmentation model is trained on that dataset and tested on the iPhone test set.

\begin{table}[]
\caption{Per-class performance of a model trained on the \ours dataset compared to a model trained on a Diffusion model dataset. Performance is expressed in mAP.}
\label{tab:diffusion_results}
\centering
\begin{tabularx}{0.7\textwidth}{Xcc}
\toprule
\textbf{Class} & \multicolumn{1}{l}{\textbf{\ours}} & \multicolumn{1}{l}{\textbf{Diffusion Model}} \\
\midrule
Bottle         & \textbf{88.44}                             & 66.67                                        \\
Eyedrops       & \textbf{79.39}                             & 51.67                                        \\
Plant          & \textbf{69.89}                             & 4.72                                         \\
Sprite         & \textbf{93.07}                             & 62.40                                        \\
Vase           & \textbf{75.29}                             & 53.20                                        \\
\midrule
Total          & \textbf{81.21}                             & 48.86 \\
\bottomrule
\end{tabularx}
\end{table}

From the results in \cref{tab:diffusion_results}, we observe that our approach also outperforms the Diffusion model-based data generation by a large margin. This holds true for all classes. Although the images generated by the Diffusion model look very realistic, the model sometimes significantly deforms the objects to make them fit better in the background. This can warp the representation of the downstream model for those objects, causing worse performance. Data generation approaches based on Diffusion models that are specifically designed for synthetic data should lead to better performance.

\section{Future Work and Limitations}

There still are limitations to our approach that could inspire future work. Some of these limitations are inherent to the current state of Gaussian Splatting. These models do not support complex lighting features, such as relighting and refractions. Transparent objects are also often not represented correctly. In this study, we used data augmentation to simulate different lighting effects. However, this is not realistic. Direct relighting of the spherical harmonics of the Gaussian Splatting model based on the background image could result in better-quality images. Additionally, the effects of the placed object on the background representation, such as shadows and ambient lighting, are not considered.

Our approach limits itself to generating synthetic data of objects standing upright on flat surfaces. While much of the demand for synthetic data falls under this category, other scenarios could benefit from synthetic data that are not supported by our method. For example, a bin-picking scenario where objects are tossed randomly in a box.

Finally, our approach does not consider the side of the object that is on the floor during the recording of the input video. Similarly, artifacts are sometimes visible in the rendered images due to the bottom of objects being removed by the plane filter. This can be an issue for objects that are not very tall or have a large contact surface with the floor. Future work can avoid this by recording the object in multiple poses and merging the Gaussian Splatting models together. Additionally, a learned point cloud segmentation approach could be used to extract the foreground object with fewer artifacts.

\section{Conclusion}

In this work, we proposed an approach for generating synthetic data that tries to overcome the limitations of current approaches. Cut-and-Paste methods are hindered by the limited amount of variation in viewpoints of the foreground object and the general lack of realism. Rendering-based approaches require an accurate textured object model and a 3D representation of the background scene. Generative AI models, on the other hand, require fine-tuning or retraining to generate specific objects.

Our approach leverages Gaussian Splatting to avoid the need for textured 3D models while introducing many variations in the representation of the target objects at a high level of realism. Our approach is very convenient as the only manual effort required is a video of the target object.

To evaluate our approach and other future image-based synthetic data generation approaches, we introduced the IBSYD dataset. From experiments on this dataset, we have concluded that instance segmentation models trained on data generated by our method achieve good performance. Additionally, we have shown that the smart placement employed by our technique leads to better-performing models as the generated data is plausible. As a benchmark, we have compared our approach to two alternative image-based synthetic data generation approaches. This comparison has shown that the added viewpoint variation and plausible object placement lead to better results compared to Cut-and-Paste. Additionally, we have shown that the consistency with the original object representation in our method gives us a significant edge over the Diffusion model-based approach.

\begin{credits}
\subsubsection{\ackname} This study was supported by the Special Research Fund (BOF20OWB24) of Hasselt University and by the FWO fellowship grant (1SHDZ24N). The research was carried out within the framework of the
NORM.AI SBO project (Natural Objects Rendering for Economic AI Models), funded by Flanders Make, the strategic research centre for the Manufacturing Industry in Belgium. This work was made possible with support from MAXVR-INFRA, a scalable and flexible infrastructure that facilitates the transition to digital-physical work environments.

\end{credits}
%
%
%
\bibliographystyle{splncs04}
\bibliography{bib}

\end{document}